\documentclass{article}

\usepackage[preprint, nonatbib]{arxiv}
\usepackage[numbers]{natbib} 

\usepackage{color}
\usepackage{algorithm}
\usepackage[utf8]{inputenc} 
\usepackage[T1]{fontenc}    
\usepackage{url}            
\usepackage{booktabs}       
\usepackage{amsfonts}       
\usepackage{nicefrac}       
\usepackage{microtype}      
\usepackage{xcolor}         
\usepackage{amsmath}
\usepackage{todonotes}
\usepackage{tikz}
\usepackage{makecell}
\usepackage{listings}
\usepackage[british,american]{babel}
\usepackage{enumitem}

\title{Normalized Convolutional Neural Network}

\usepackage{mwe}
\usepackage[export]{adjustbox}

\author{
 Dongsuk Kim$^{1}$ \quad Geonhee Lee$^{1}$  \\ \textbf{\quad  Myungjae Lee$^{2}$ \quad Shin Uk Kang$^{2}$ \quad Dongmin Kim$^{2}$} \\ 
  $^2$ JLK Group(Research)\\
  $^1$\texttt{{kds900910}@gmail.com} \\
}
\begin{document}
\maketitle
\begin{abstract}

We introduce a Normalized Convolutional Neural Layer, a novel approach to normalization in convolutional networks. Unlike conventional methods, this layer normalizes the rows of the im2col matrix during convolution, making it inherently adaptive to sliced inputs and better aligned with kernel structures. This distinctive approach differentiates it from standard normalization techniques and prevents direct integration into existing deep learning frameworks optimized for traditional convolution operations. Our method has a universal property, making it applicable to any deep learning task involving convolutional layers. By inherently normalizing within the convolution process, it serves as a convolutional adaptation of Self-Normalizing Networks, maintaining their core principles without requiring additional normalization layers. Notably, in micro-batch training scenarios, it consistently outperforms other batch-independent normalization methods. This performance boost arises from standardizing the rows of the im2col matrix, which theoretically leads to a smoother loss gradient and improved training stability.
\end{abstract}


\section{Introduction}

\indent Normalization methods have significantly enhanced the performance of deep neural networks. Among them, Batch Normalization (BN) \cite{BatchNorm} is widely adopted in deep learning architectures due to its ability to accelerate training and help models converge to a better minimum of the loss function \cite{HowBN}. However, BN is particularly effective in large-batch environments, as it is designed to mitigate internal covariate shift.
To address BN’s limitations with small batch sizes, various batch-independent normalization methods and micro-batch networks have been proposed. Nevertheless, these methods still struggle to match the performance of BN when applied to large-batch training.
In this paper, we introduce the Normalized Convolutional Neural Network (NC Neural Network), a novel approach with greater generality across various convolutional neural network (CNN) tasks. Similar to BN, NC smooths the loss landscape by standardizing sliced inputs in response to convolutional filters. However, unlike conventional normalization methods, which are applied before or after convolution, our approach directly normalizes the rows of the im2col matrix during the convolution process.
This fundamental shift means that NC operates within the convolution process itself, making it distinct from existing normalization techniques. As a result, NC exhibits a universal property, allowing it to be applied to diverse CNN-based tasks, including object detection, instance segmentation, and more.
\indent To demonstrate the effectiveness of NC, we provide both theoretical and empirical analyses. Our key contributions are:

\begin{itemize}
\item \textbf{Theoretical Insight.} Theoretical Insight: We show that NC reduces the Lipschitz constants of the loss and gradient functions, leading to a smoother loss landscape in deep convolutional networks.

\item \textbf{Empirical Validation.} Experimental results indicate that networks using NC outperform those using Group Normalization (GN) \cite{GroupNorm} in image classification, object detection, and segmentation.

\item \textbf{Generative Model Enhancement.} We demonstrate that NC provides significant improvements in generative models compared to Positional Normalization.
To further establish NC’s generality across vision tasks, we conduct extensive experiments, including:

Image classification on multiple datasets,
Object detection and instance segmentation on the COCO dataset \cite{COCO},
Semantic segmentation on PASCAL VOC \cite{PSACALVOC}, and
Image generation on various benchmark datasets \cite{CGAN}.

\end{itemize}

\section{Related Works}

Normalization methods have been widely used since the early days of deep learning. For instance, Local Response Normalization (LRN) was employed in AlexNet \cite{ALEX}. Batch Normalization (BN) was later introduced to mitigate internal covariate shift, leading to improved training stability and convergence speed. BN is particularly effective for large batch sizes, but its performance tends to decline when the batch size is reduced.

To address this limitation, various normalization techniques have been proposed that do not rely on the batch dimension. Layer Normalization (LN) normalizes features along the channel dimension, while Instance Normalization (IN) applies normalization independently to each sample. Among these, Group Normalization (GN) \cite{GroupNorm} has been shown to be highly effective for micro-batch sizes. GN generalizes normalization by dividing channels into $G$ groups and applying layer normalization within each group. Positional Normalization (PONO) \cite{PN} normalizes features at each spatial location independently across channels, capturing coarser structural information of input images compared to other normalization techniques. In addition to these, numerous other normalization methods continue to be developed.

Most existing normalization techniques can be formulated as follows. Given an input tensor $X \in \mathbb{R}^{B \times C \times H \times W}$, where $B$ denotes the batch size, $C$ the number of channels, and $H \times W$ the spatial dimensions, normalization is applied along specific axes. The general formulation is:
\begin{equation*}
X'_{K} = \beta \frac{X_{K}- \mu_{X_{K}}}{\sigma_{X_{K}}} + \gamma,
\end{equation*}
where $K$ represents the axes along which normalization is performed. The learnable parameters $\beta$ and $\gamma$ are applied along the remaining axes. For example, BN normalizes across $K={B, H, W}$ and learns $\beta$ and $\gamma$ in $\mathbb{R}^C$. Based on this formulation, various techniques have emerged that exploit different strategies for $\beta$ and $\gamma$. Notably, methods that leverage style information, such as Conditional Instance Normalization (CIN) and Adaptive Instance Normalization (AdaIN) \cite{AdaIN,CIN}, have been explored. PONO further utilizes a technique called Moment Shortcut to effectively capture structural information \cite{PN}.

Normalization techniques are known to smooth non-convex loss landscapes, thereby enhancing training stability and convergence. We emphasize that NC exhibits a similar effect, reducing non-convexity while preserving structural information by standardizing regions corresponding to the convolutional kernel.

\section{Normalized Convolutional Networks}
\label{sec:headings}

Batch Normalization (BN) fundamentally influences network training by significantly smoothing the loss landscape \cite{HowBN}, leading to a more stable and faster training process. This impact extends to the recently introduced normalization method, Weight Standardization (WS) \cite{WS}. While BN primarily considers the Lipschitz constants of activations, WS focuses on the weights. Since the gradients with respect to the weights also depend on the inputs interacting with these weights, WS also contributes to smoothing the loss landscape.

Building on this idea, we propose that normalizing each sliced input corresponding to a convolutional kernel can similarly smooth the loss landscape. Thus, we argue that NC can be regarded as the dual of WS.

\subsection{Normalized Convolution}
Consider a standard convolutional layer with bias terms set to zero:

\begin{equation*}
    y=W*x
\end{equation*}
where ${W\in R^{O \times I}}$ denotes the weights and ${x\in R^{I \times HW}}$ denotes the input in the layer. The symbol $\ast$ represents the convolution operation, which, in practice, corresponds to matrix multiplication. For ${W \in R^{O \times I}}$, ${O}$ represents the number of output channels, while ${I}$ to the number of input channels within the kernel region of each output channel. Similarly, for ${x \in R^{I \times HW}}$, the term ${HW}$ denotes the prodcut of output width and output heights. NC normalizes the im2col matrix ${x}$ as follows:

\begin{equation*}
    \widehat{x}_{i,k} = \frac{x_{i,k} - \mu_{k}}{\sigma_{k} + \epsilon},
\end{equation*}
where

$$\centering \mu_{k} = \frac{1}{I} \sum_{i=1}^I x_{i,k}$$ $$\sigma_{k}=
 \sqrt{\frac{1}{I} \sum_{i=1}^I(x_{i,k}-\mu_{k})^2}.$$

After normalizing the im2col matrix, the output can be computed via matrix multiplication. Notably, the affine transformation can be applied either before or after the convolution operation. In our experiments, we apply the affine transformation after convolution.

\begin{figure}[t] 
    \centering
    \includegraphics[scale=0.3]{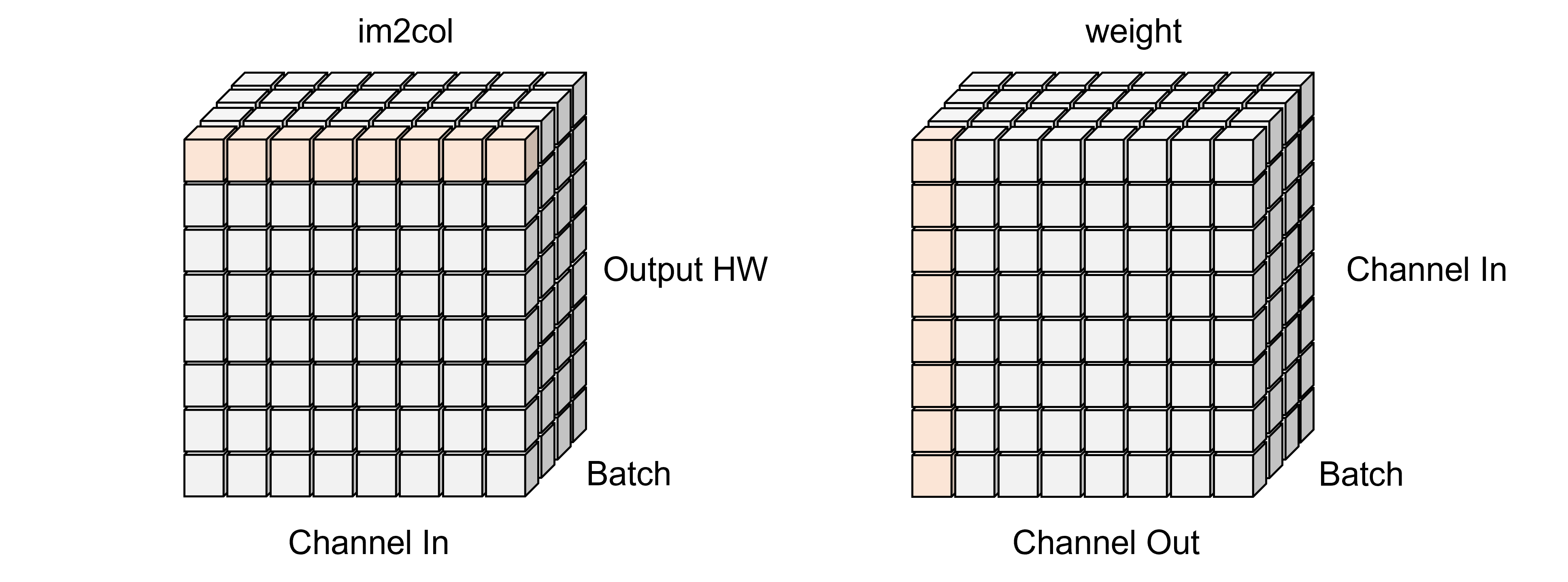}
    \caption{Normalized convolutional layer diagram. The rows of im2col matrix in the pink area are stadardized and the correspond area of weights can be also stadardized.}
    \label{fig:fig1}
\end{figure}
\subsection{Gradients of Normalized Convolution}

We will now demonstrate that NC helps smooth the loss landscape. The calculations involved are closely related to those in \cite{WS} and are explained in greater detail in that paper. Specifically, we show that optimizing $L$ with respect to $x$ has smaller Lipschitz constants on both the loss function and the gradients than optimizing $L$ on ${\widehat{x}}$. Lipschitz constant of a function $f$ is any value $A$ that satisfies: 
\begin{equation*}
    {\left| {f} \left( x_{1} \right) - {f} \left( {x_{2}} \right) \right| } {\le {A} \parallel{x_{1} - x_{2}}\parallel}
\end{equation*} for all $x_1, x_2$.

For a neural network, $f$ corresponds to the loss function $L$ or its gradient ${\nabla_{x}{L}}$. Smaller Lipschitz constants for the loss and its gradients imply that changes in the loss and gradients during training are more controlled, allowing the optimizer to take larger steps and thereby accelerating learning.

The normalization process consists of two steps. The first step centers each row of the im2col matrix by subtracting its mean, and the second step normalizes by dividing by the standard deviation. We rewrite this process in vector form as follows: 

\begin{equation*}
\dot{x_{.,k}} = x_{.,k} - \frac{1}{I}\mathbf{1}<\mathbf{1},x_{.,k}>
\end{equation*}

\begin{equation*}
\widehat{x}_{.,k} = \dot{x_{.,k}} / (\sqrt{\frac{1}{I}<\mathbf{1},\dot{x_{.,k}}^{\circ^2}>})
\end{equation*}
where $< , >$ denotes dot product and $\circ ^2$ represents Hadamard product.

The corresponding gradient vectors are:

\begin{equation*}
\nabla_{\dot{x_{.,k}}}{L} = \frac{1}{\sigma_{x_{k}}}(\nabla_{\widehat{x}_{.,k}}{L}-\frac{1}{I}<\widehat{x}_{.,k},\nabla_{\widehat{x}_{.,k}}{L}> \widehat{x}_{.,k})
\end{equation*}

\begin{equation*}
\nabla_{x_{.,k}}L = \nabla_{\dot{x_{.,k}}}{L}-\frac{1}{I}\mathbf{1}<\mathbf{1},\nabla_{\dot{x_{.,k}}}{L}>
\end{equation*}

To analyze the Lipschitz constant of the loss, a straightforward calculation yields:
\begin{equation}
{\parallel{\nabla_{\dot{x_{.,k}}}}{L}\parallel}^2 = \frac{1}{\sigma_{k}}\left({\parallel{\nabla_{\widehat{x}_{.,k}}{L}\parallel}^2+\frac{1}{I^2}<\widehat{x}_{.,k}},\nabla_{\widehat{x}_{.,k}}{L}>^2\left(<\widehat{x}_{.,k},\widehat{x}_{.,k}>-2I\right) \right)
\end{equation}. Due to the standardization process, we have ${{\parallel{\widehat{x}_{.,k}}\parallel}^2 = I}$.  Since NC is a convolutional process followed by normalizing each row of the im2col matrix corresponding to $k$, the effect of ${ \frac{1}{{\sigma_{x}}_{k}} }$ is neutralized. Consequently, the primary effect on the gradient norm is the reduction term: ${\frac{1}{I}<{\widehat{x}_{.,k}},\nabla_{\widehat{x}_{.,k}}{L}>^2}$.

Next, similar calcuation shows,
\begin{equation}
{\parallel{\nabla_{{x}_{.,k}}}{L}\parallel}^2 ={\parallel{\nabla_{\dot{{x}_{.,k}}}}{L}\parallel}^2 -  \frac{1}{I}<1,\nabla_{\dot{x}_{.,k}}{L}>^2
\end{equation}

Thus, the effect of (2) on the Lipschitz constant is also a reduction.

It is important to note that the gradient of the loss with respect to the 
$k$-th layer is obtained by matrix multiplication of the gradient of the loss with respect to the inputs and the gradient of the activation function of the 
$k$-th layer with respect to its input variables. Therefore, understanding how the loss gradient behaves with respect to input variables is crucial.
Summarizing (1) and (2), we can say NC make loss landscape smoother. To verify this effect on real world, we conducted case study on ResNet-50 \cite{resnet} trained on ImageNet \cite{ImageNet} and ResNet-18 on CIFAR-10/100. After that, Experiments were also conducted on object detection, semantic segmentation, and image generation tasks on several famous dataset.

Summarizing (1) and (2), we conclude that NC helps smooth the loss landscape. To validate this effect in real-world settings, we conducted case studies on ResNet-50 \cite{resnet} trained on ImageNet \cite{ImageNet} and ResNet-18 trained on CIFAR-10/100. Additional experiments were conducted on object detection, semantic segmentation, and image generation tasks across several well-known datasets.

\subsection{CUDA Kernel Implementation}

The pseudo-code using PyTorch is presented in Algorithm 1. Running the experiment with existing PyTorch operations, such as unfold for manually constructing an im2col matrix, results in considerable slowdowns. To address this, I developed a custom CUDA kernel. The implementation is available on GitHub: \footnote{\url{https://github.com/kimdongsuk1/NormalizedCNN.}}

\begin{algorithm}[t]
\caption{Normalized Convolutional Layer Pseudocode, PyTorch-like}
\label{alg:code}
\definecolor{codeblue}{rgb}{0.25,0.5,0.5}
\definecolor{codekw}{rgb}{0.85, 0.18, 0.50}
\lstset{
  backgroundcolor=\color{white},
  basicstyle=\fontsize{7.5pt}{7.5pt}\ttfamily\selectfont,
  columns=fullflexible,
  breaklines=true,
  captionpos=b,
  commentstyle=\fontsize{7.5pt}{7.5pt}\color{codeblue},
  keywordstyle=\fontsize{7.5pt}{7.5pt}\color{codekw},
}
\begin{lstlisting}[language=python]
def NCon2d(inputs, in_channel, out_channel, kernel_size): 

    h,w = output_height,output_width
    flatten_weights = flatten(weight,out_channel,kernel_size)
    
    im2col_inputs = unfold(x,kernel_size,channel)
    mean = im2col_inputs.mean(axis=-1)
    std = (im2col_inputs.var(axis=-1)+epsilon).sqrt()
    im2col_inputs = (im2col_inputs - mean) / std
    
    output = im2col_inputs @ flatten_weights.
    
    return output.reshape(-1,h,w,out_channel)
\end{lstlisting}
\end{algorithm}

\section{Experiment}

\subsection{Image Classification on ImageNet, CIFAR-10 and 100}

\begin{table}[t]
\caption{Results of Experiments on image classification tasks}
\centering
\small
\begin{tabular*}{\textwidth}{c @{\extracolsep{\fill}} lccc}
\hline
Dataset   &Model    &Method  & Top-1  \\ 
\hline
ImageNet       &ResNet-50 &GN & 24.95      \\ 
& &NC & \textbf{24.24}    \\
& &GN+NC & \textbf{23.81}     \\
\hline
CIFAR-10  &ResNet-18&GN & 10.14       \\

&&NC & \textbf{8.52}       \\

&&GN+NC & \textbf{6.98}       \\
\hline
CIFAR-100            &ResNet-18&GN & 34.65      \\
&  & NC & \textbf{27.20}       \\
&  & GN+NC & \textbf{26.15}        \\
\hline
\end{tabular*}
\label{tab:table1}
\end{table}

\begin{figure}[tp] 
    \centering
    \includegraphics[scale=0.4]{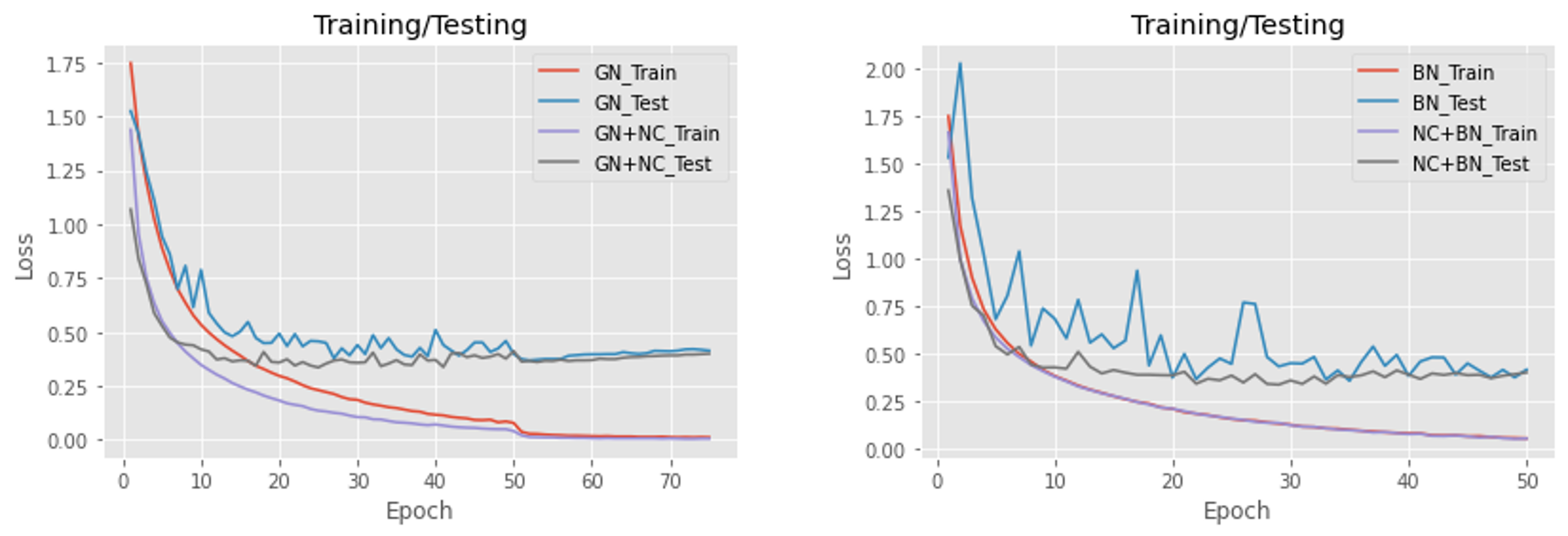}
    \caption{The graph illustrates the train and test loss values during training and validation on CIFAR-10. It can be observed that the training process is more stable, and the loss value decreases at a faster rate.}
    \label{fig:fig2}
\end{figure}

The ImageNet dataset is a large-scale image dataset consisting of approximately 1.28 million training samples and 50,000 validation images, spanning 1,000 categories. We conduct a simple comparison between ResNet-50 with Group Normalization (GN) and ResNet-50 with NC in a micro-batch setting, following the standard training setup for ImageNet. Since NC employs a batch-independent normalization process, we compare it specifically with Group Normalization, which has been shown to outperform other batch-independent normalization methods such as Layer Normalization and Instance Normalization. For all models, the number of groups in GN is set to 32.

All experiments are conducted using PyTorch implementations, with NC implemented as a custom layer in PyTorch. The training batch size for all models is set to 2.

Although the CIFAR-10 dataset is smaller than ImageNet, it is one of the most widely used datasets in machine learning research. The CIFAR-10/100 dataset contains 60,000 32×32 color images across 10 and 100 different classes, respectively. We conduct the same experiment on CIFAR-10/100 using ResNet-18. In this case, we use vanilla SGD without momentum and weight decay. The training batch size is again set to 2, and all models use ReLU activation. We apply basic data augmentations, including horizontal flipping and random shifting of up to 10$\%$ of the image size. All models are trained for 75 epochs.

Additionally, we conduct experiments to analyze the effect of combining NC with existing normalization methods. Specifically, we compare ResNet-18 with BN/GN against ResNet-18 with BN/GN and NC. While the final accuracy and validation loss values in the batch normalization cases show no significant difference, we observe a faster learning speed during the initial training phase, as expected. However, in the micro-batch setting, we observe more notable improvements in accuracy. See Figure 2 and Table 1 for details.

Across all experiments, NC consistently outperforms GN. Since NC is essentially a convolutional layer, we also evaluate its performance in combination with other normalization techniques. The results show that NC further improves performance when used alongside two representative normalization methods, BN and GN.

\subsection{Object Detection and Semantic Segmentations}

Micro-batch training is commonly used in object detection and segmentation tasks. We conduct experiments using pre-trained ResNet-50 models with GN and NC on COCO-2017 for object detection and PASCAL-VOC 2012 for semantic segmentation. In all experiments, the batch size is set to 2 images per GPU, and transfer learning is applied.

For object detection, we employ Faster R-CNN, following the standard PyTorch-based Mask R-CNN framework \footnote{\url{https://github.com/facebookresearch/maskrcnn-benchmark}}. The model is equipped with a Feature Pyramid Network (FPN), a 4-conv-1fc bounding box head, and a 1× learning rate schedule. We evaluate the model's performance using Average Precision (AP), AP$^{0.50}$, and AP$^{0.75}$ for bounding box detection.

For semantic segmentation, we use DeepLabV3 with the same pre-trained ResNet-50 backbone. The evaluation is conducted on 21 different classes, following the well-established experimental setup in \footnote{\url{https://github.com/VainF/DeepLabV3Plus-Pytorch}}. The batch size is set to 2, and model performance is assessed using the Mean Intersection over Union (mIoU) metric.

As shown in Table 3, the results in both tasks exhibit trends similar to those observed in image classification, further confirming the benefits of NC in combination with existing normalization methods. These findings suggest that NC enhances performance in well-known convolution-based tasks. Future work will focus on verifying these effects across various backbone architectures and extending the analysis to instance segmentation.

\begin{table}[t]
\caption{Results of Experiments on object detection and semantic segmentation}
\centering
\begin{tabular}{l|c|c|c|c|c|c}
\hline
Tasks   & DataSet  & Model     & AP & AP$^{0.5}$ & AP$^{0.75}$ & mIoU \\
\hline
Detection & COCO & GN & 38.01     & 59.12 & 41.17     & -\\
 &  & NC   & 38.59     & 59.91 & 41.55 &-\\
 &  & GN+NC   & 39.01     & 60.34 & 42.07 &-\\
\hline
Segementation & PASCAL& GN & -    & -     & -& 74.33\\
 &  & NC   & - & - & - &75.90\\
 &  & GN+NC   & - & - & - &76.98\\

\end{tabular}
\label{tab:table2}
\end{table}

\begin{table}[t]
\caption{Results of Experiments on bi-directional image generation using CycleGAN and Pix2pix}
\centering
\scalebox{0.8}{
\begin{tabular}{l|c|c|c|c|c|c}
\hline
Dataset   &Baseline     & Method     & FID (A $\to$ B) & FID (B $\to$ A) & LPIPS (A $\to$ B) & LPIPS (B $\to$ A)\\
\hline
Maps &CycleGAN & PONO+MS & 52.45 & 64.13     & 0.3317 & \textbf{0.175} \\
 &  & NC+MS   & \textbf{49.57} & \textbf{59.78}     & \textbf{0.327} & 0.177 \\
\hline
Horse2Zebra & CycleGAN & PONO+MS & 71.70    & 136.61 & 0.757     & \textbf{0.730} \\
 &  & NC+MS   & \textbf{67.54}   & \textbf{133.81} & \textbf{0.731}     & 0.763 \\

\hline
Maps &Pix2pix & PONO+MS & 57.39     & 62.39 & 0.334 & 0.171\\
 &  & NC+MS   & \textbf{52.85} & \textbf{60.01} & \textbf{0.329} & 0.171\\
\hline
Day2Night &Pix2pix & PONO+MS & 200.69      & 188.49   & \textbf{0.6198}  & 0.5781 \\
 &  & NC+MS   & \textbf{195.01}      & \textbf{179.97}& 0.624     & \textbf{0.5356} \\

\hline
\end{tabular}}
\label{tab:table2}
\end{table}

\begin{figure}[tp] 
    \centering
    \includegraphics[scale=0.4]{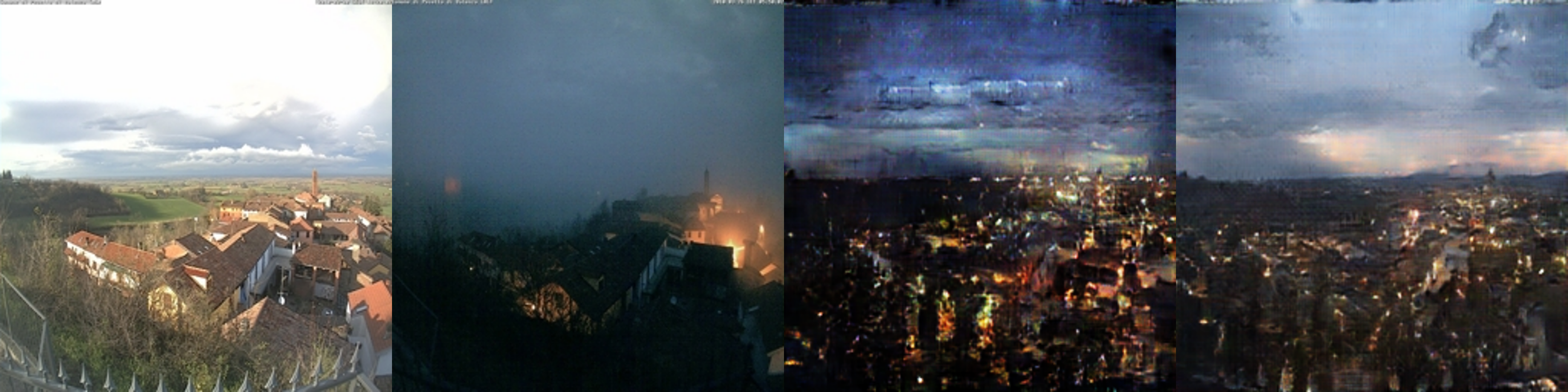}
    \caption{A pair of test images is presented. The first two images are the original images, the third image is generated using PONO, and the last image is produced using NC.}
    \label{fig:fig2}
\end{figure}

\subsection{Comparison Structural Effect on Generative Models}

\textbf{Relation to Positional Normalization} NC is equivalent to Positional Normalization (PN) when the corresponding kernel size is 1×1 in a 2D convolutional process. Thus, it is natural to compare NC with PN. Notably, NC is designed to better preserve structural information, as it normalizes the channels corresponding to the weights, ensuring that key spatial features remain intact.

\textbf{Deep Generative Models} Most deep generative models employ an encoder-decoder architecture. Since the introduction of Generative Adversarial Networks (GANs) \cite{GAN}, numerous variants have been developed, achieving remarkable results. For instance, Zhu et al. \cite{CGAN} introduced image-to-image translation between input and output images, while Huang et al. \cite{Pix} enabled style transfer from an input image A to an output image B.
U-Net \cite{UN}, a well-known encoder-decoder model, has demonstrated outstanding performance in segmentation tasks. The core principles of U-Net have also been effectively applied to various generative models.

\textbf{Moment Shortcut} Positional Normalization (PONO) captures structural signature information at different layers. Leveraging this idea, the Moment Shortcut (MS) technique has been proposed. MS forwards the positional moment information 
$\mu$ and $\sigma$, extracted from the encoder layers, to the deconvolutional decoder layers. This technique can be applied independently of normalization.

To compare structural feature extraction capabilities, we evaluate NC (+MS) and PONO (+MS) using two baseline models: CycleGAN and Pix2Pix. The only difference lies in whether normalization or NC is used. For all models, we follow the same experimental setup using the official implementations provided in \footnote{\url{https://github.com/junyanz/pytorch-CycleGAN-and-pix2pix}}. Additionally, we employ concatenated skip connections, where encoder activations are directly concatenated with decoder activations, similar to U-Net. For further architectural details, refer to \cite{PN}.

\textbf{Evaluation Metrics} We use two widely adopted evaluation metrics:

\begin{itemize} 
\item \textbf{Fréchet Inception Distance (FID) \cite{FID}}: This metric computes the distance between the distributions of generated images and real test images in the target domain. It is based on the Gaussian distribution of extracted features from a pre-trained Inception-V3 model trained on ImageNet. FID provides an estimate of how closely the generated images match the real data distribution.

\item \textbf{Learned Perceptual Image Patch Similarity (LPIPS) \cite{LPIPS}}: This metric measures the perceptual similarity between pairs of generated and target images. LPIPS is computed using features extracted from AlexNet \cite{ALEX}. It has been shown \cite{LPIPS} to strongly correlate with human perceptual judgments of image similarity.

\end{itemize}

\textbf{Dataset} We conduct experiments on three datasets:

\begin{itemize} 
\item \textbf{Horse2Zebra} This dataset consists of 1,067 horse images and 1,334 zebra images for training, along with 120 horse images and 140 zebra images for testing. The dataset is unpaired and derived from ImageNet.
\item \textbf{Maps \cite{CGAN}} This dataset includes 1,069 training images and 1,098 testing images for each domain.
\item \textbf{Day2Night \cite{Day}} This dataset contains 17,823 natural scene images for training and 2,287 images for testing.
\end{itemize}

Following the official CycleGAN and Pix2Pix implementations, we integrate and replace PN with NC. The first two datasets are used for CycleGAN, while the second and third datasets are used for Pix2Pix.

All models are trained for 200 epochs, and FID and LPIPS are computed on the final results. When evaluating NC, we replace only the standard convolutional layer with our NC layer while maintaining the same layer positions as in PONO. Across all experiments, NC yields performance improvements over existing methods. The results, presented in Table 3, confirm that NC enhances generative model performance. Additionally, qualitative improvements can be observed in Figures 3 and 4.

\begin{figure} 
    \centering
    \includegraphics[scale=0.6]{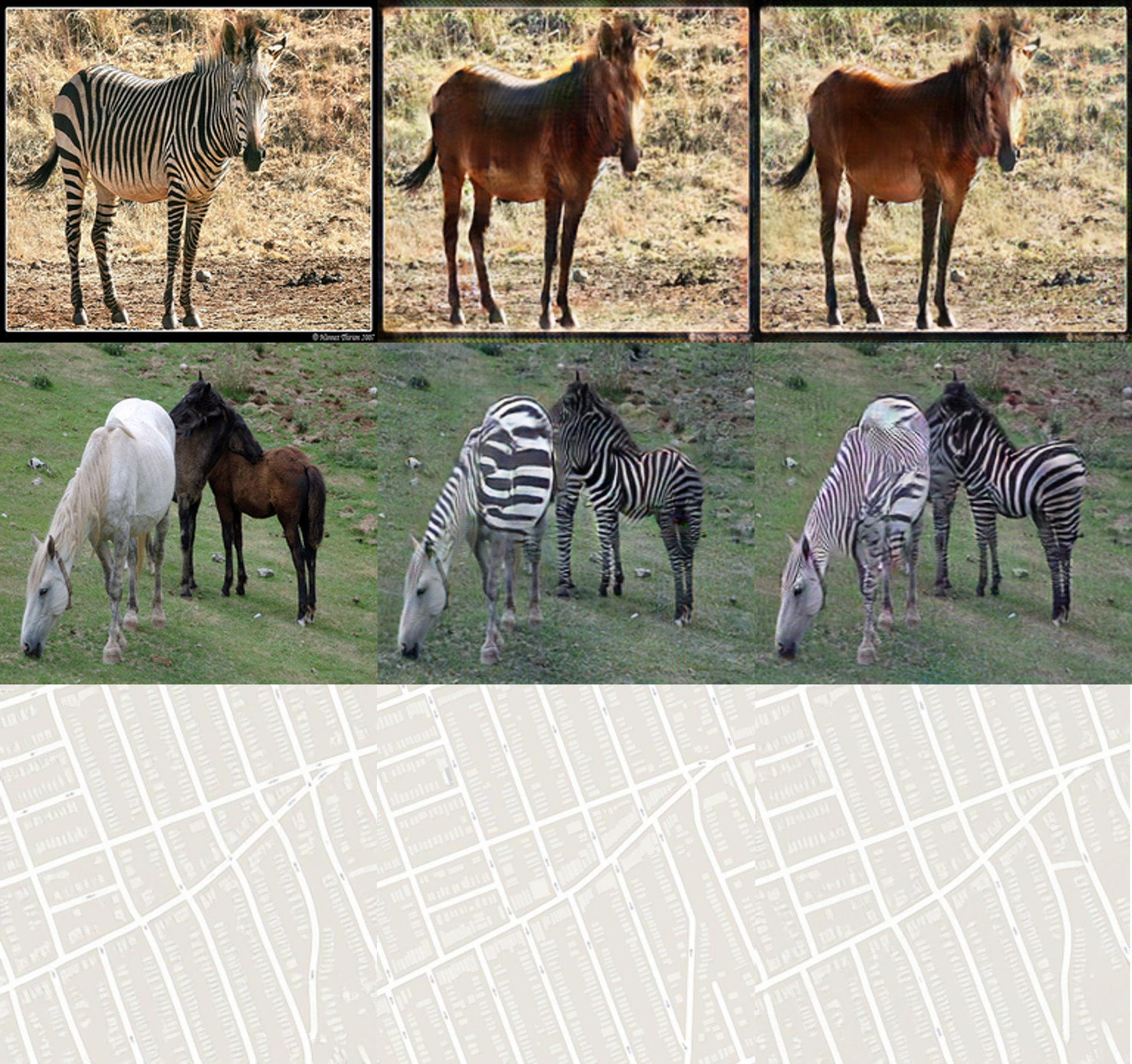}
    \caption{The images were obtained using CycleGAN. The first image represents the original, the second image is generated using PONO, and the third image is produced using NC. Notably, the image obtained from NC contains more detailed information, which can be intuitively observed.}
    \label{fig:fig2}
\end{figure}

\section{Conclusion and Future work}

In this paper, we introduce Normalized Convolution (NC), a novel convolution paradigm inspired by the smoothing effects of Weight Standardization (WS) \cite{WS} during training. Our approach can be considered a dual version of WS, aligning with the fundamental properties of differential convolution operators. Specifically, if one function in a convolutional operation is smooth, the overall convolution tends to inherit the smoothness properties of that function, even if the other function is not inherently smooth. Through theoretical analysis, following the calculations in \cite{WS}, we demonstrate that NC also reduces the Lipschitz constant, further contributing to training stability.

Empirical results indicate that NC outperforms Group Normalization (GN), making it a strong candidate for micro-batch image classification and related tasks. Based on the findings in \cite{WS}, we hypothesize that WS+NC could serve as a state-of-the-art method in this setting. However, our experiments did not yield significant improvements when combining WS with NC, suggesting that further investigation is required.

Among batch-independent normalization techniques, Positional Normalization (PN) \cite{PN} shares similarities with NC, as both methods coincide when applied with a 1×1 kernel. However, since the kernel size is not always 1×1, NC provides a more adaptive approach by directly normalizing slice-inputs through the im2col matrix. Our experiments indicate that PN is more suited for GAN-based tasks, whereas NC demonstrates improvements when applied to these models. Further investigation is needed to fully explore NC’s potential in various tasks.

Another key direction for future research is identifying a more suitable activation function for NC. While SELU \cite{SELU} exhibits self-normalizing properties, our experiments suggest that it is not an optimal choice for NC. With an improved initialization strategy, SELU may still prove beneficial, and future work should explore this possibility.

Finally, we observed that NC does not perform well with adaptive gradient-based optimizers such as Adam \cite{Adam}. This suggests that NC requires a specialized adaptive optimization method to fully leverage its benefits. Future research should focus on developing an optimization strategy tailored to NC, potentially improving training dynamics and overall model performance.

In summary, our study highlights NC as a promising normalization technique with applications in various domains. However, further research is necessary to refine its integration with different architectures, optimizers, and activation functions, paving the way for more robust and effective deep learning models.

\section{Acknowledgments}
This research was financially supported by the Ministry of Trade,Industry, and Energy(MOTIE), Korea, under "Regional Specialized Industry Development Program(R\&D,P0002072)" supervised by the Korea Institute for Advacement of Technology(KIAT).

\bibliographystyle{plain}
\bibliography{z}

\end{document}